\pdfoutput=1
\documentclass[11pt]{article}

\usepackage[final]{acl}

\usepackage{times}
\usepackage{latexsym}

\usepackage[T1]{fontenc}
\usepackage[utf8]{inputenc}

\usepackage{microtype}

\usepackage{inconsolata}

\usepackage{graphicx}
\usepackage{amsmath}
\usepackage[symbol]{footmisc}
\usepackage{csquotes}
\usepackage{epigraph}
\usepackage{listings}
\usepackage{hyperref}

\usepackage{enumitem}

\usepackage{amssymb}% http://ctan.org/pkg/amssymb
\usepackage{pifont}% http://ctan.org/pkg/pifont
\newcommand{\cmark}{\ding{51}}%
\newcommand{\xmark}{\ding{55}}%

\usepackage{algorithm}
\usepackage{algpseudocode}

\usepackage{multirow}
\title{The Base-Rate Effect on LLM Benchmark Performance: \\ {\large Disambiguating Test-Taking Strategies from Benchmark Performance}}

\author{Kyle Moore\textsuperscript{\textdagger}\\
  Vanderbilt University \\
  \texttt{kyle.a.moore@vanderbilt.edu} 
  \\\And
  Jesse Roberts\textsuperscript{\textdagger} \\
  Tennessee Tech University \\
  Vanderbilt University \\
  \texttt{jtroberts@tntech.edu} 
  \\ \AND
  Thao Pham \\
  Berea College \\
  \\ \And
  Oseremhen Ewaleifoh \\
  Vanderbilt University \\
  \\ \And
  Doug Fisher \\
  Vanderbilt University \\
}

\setlist{nosep}

\begin{document}
\maketitle
\begin{abstract}

Cloze testing is a common method for measuring the behavior of large language models on a number of benchmark tasks. Using the MMLU dataset, we show that the base-rate probability (BRP) differences across answer tokens are significant and affect task performance ie. guess \textit{A} if uncertain. We find that counterfactual prompting does sufficiently mitigate the BRP effect. The BRP effect is found to have a similar effect to test taking strategies employed by humans leading to the conflation of task performance and test-taking ability. We propose the Nvr-X-MMLU task, a variation of MMLU, which helps to disambiguate test-taking ability from task performance and reports the latter.

\end{abstract}

\section{Introduction}
    \footnotetext[2]{Equal Contribution}
    \footnotetext[1]{https://github.com/KyleAMoore/MMLU-cloze-vs-cf}
    
    % \epigraph{
    %     \textit{When a measure becomes a target, it ceases to be a good measure} - Charles Goodhart
    % }
    
    Benchmarking has become an ubiquitous practice in Machine Learning. Ideally, these benchmarks provide human-interpretable measures of context specific abilities \cite{storks2019recent}, however, as standardized tests, benchmarks may be susceptible to response strategies that skew the reported metrics and belie their utility \cite{cordon1996strategy}. 
    
    In the context of large language models (LLMs), many benchmarks are measured by way of cloze testing \cite{storks2019recent}, with the most probable allowed completion considered to be the model's preferred answer. We suspect that the identified preference of many models may be undesirably biased by the independent intrinsic probabilities associated with each completion option, referred to as the base-rate probability (BRP) effect.

    To address the formal hypotheses in Table \ref{tab:hyp} we: (1) quantify the BRP effect on model accuracy on the Massive Multitask Language Understanding (MMLU) task \cite{hendrycks2020measuring} (H1\cmark) and find that accuracy is strongly affected by correct answer label (H2\cmark); consider (2) the BRP effect when counterfactual prompting is used to measure preference and find that the effect is mitigated but remains (H3\xmark); finally, (3) propose a novel variation of the MMLU referred to as Nvr-X-MMLU that mitigates the BRP effect and permits a more meaningful measure of model performance (H4\cmark). 

    % \vskip4pt
    % \begin{itemize}[topsep=1pt]
    %     \small 
    %     \item[\textbf{H1}] \textit{The BRP density of answer choice tokens is not evenly distributed}
    %     \item[\textbf{H2}] \textit{BRP disparities influence cloze test answer selection}
    %     \item[\textbf{H3}] \textit{A proposed alternative, counterfactual prompting, mitigates the BRP effect on answer choice selection}
    %     \item[\textbf{H4}] \textit{Benchmark task variations can disambiguate BRP effects from task performance}
    % \end{itemize}
    % \vskip4pt
    \begin{table}[h!]
    \caption{We evaluate the following hypotheses.}
    \label{tab:hyp}
    \begin{tabular}{l c}
    \hline
    Hypothesis & Status \\ \hline
    \begin{tabular}[c]{@{}l@{}}
        \small \textit{\textbf{H1}: The BRP density of answer choice tokens}\\
        \small \textit{is not evenly distributed}\end{tabular}
        & \cmark  \\ 
        \hline
    \begin{tabular}[c]{@{}l@{}}
        \small \textit{\textbf{H2}: BRP disparities influence cloze test}\\
        \small \textit{answer selection}\end{tabular}
        & \cmark  \\ 
        \hline
    \begin{tabular}[c]{@{}l@{}}
        \small \textit{\textbf{H3}: A proposed alternative, counterfactual}\\
        \small  \textit{prompting, mitigates the BRP effect on}\\
        \small \textit{answer choice selection}\end{tabular}
        & \xmark  \\ 
        \hline
    \begin{tabular}[c]{@{}l@{}}
        \small \textit{\textbf{H4}: Benchmark task variations can disam-}\\
        \small \textit{biguate BRP effects from task performance}\end{tabular}
        & \cmark  \\ 
        \hline
    \end{tabular}
    \end{table}

    % URL-WITHHELD
    % \url{https://github.com/KyleAMoore/MMLU-cloze-vs-counterfactual}
    % \begin{figure}[t]
    % \centering
    % \includegraphics[width=\linewidth]{Images/Image2.png} % Reduce the figure size so that it is slightly narrower than the column. Don't use precise values for figure width.This setup will avoid overfull boxes.
    % \caption{Correct answer distributions for all subjects}
    % \label{fig:answers}
    % \end{figure}

\section{Background \& Related Work}

    The MMLU benchmark aims to jointly measure language understanding and knowledge retrieval abilities. It consists of 15908 multiple choice questions distributed across 57 subject areas. Each question is associated with four answer choices that are unevenly distributed across subjects: \textit{A} ($\mu$=0.231, $\sigma$=0.042), \textit{B} ($\mu$=0.245, $\sigma$=0.042), \textit{C} ($\mu$=0.254, $\sigma$=0.044), \textit{D} ($\mu$=0.270, $\sigma$=0.078). Models are evaluated on their accuracy at selecting the correct answer choice. However the method of selection is not prescribed, with many models reporting 0-shot and/or 5-shot performance as measured by a cloze test prompting methods.

% Please add the following required packages to your document preamble:
% \usepackage{multirow}
\begin{table*}[t]
\centering
\begin{tabular}{|ccc|}
\hline
\multicolumn{3}{|c|}{\textbf{Example Shared Context}}                                                                       \\
\multicolumn{3}{|c|}{\begin{tabular}[c]{@{}c@{}}What element is most common among the Jovian Planets?\\ (A) Hydrogen (B) Helium (C) Carbon (D) Oxygen.\\ Of the answer choices above,\end{tabular}} \\ \hline
\multicolumn{1}{|c|}{\textbf{\begin{tabular}[c]{@{}c@{}}Prompting Method\end{tabular}}} &
  \multicolumn{1}{c|}{\textbf{\begin{tabular}[c]{@{}c@{}}Method-Specific Context(s)\end{tabular}}} &
  \textbf{\begin{tabular}[c]{@{}c@{}}Token(s) Measured\end{tabular}} \\ \hline
\multicolumn{1}{|c|}{\multirow{4}{*}{\textbf{Cloze}}} &
  \multicolumn{1}{c|}{\multirow{4}{*}{\begin{tabular}[c]{@{}c@{}}the best answer choice is ( \_\_\_\_\end{tabular}}} &
  \textit{A} \\ \cline{3-3} 
\multicolumn{1}{|c|}{} & \multicolumn{1}{c|}{}                                                                 & \textit{B} \\ \cline{3-3} 
\multicolumn{1}{|c|}{} & \multicolumn{1}{c|}{}                                                                 & \textit{C} \\ \cline{3-3} 
\multicolumn{1}{|c|}{} & \multicolumn{1}{c|}{}                                                                 & \textit{D} \\ \hline
\multicolumn{1}{|c|}{\multirow{4}{*}{\textbf{CF}}} &
  \multicolumn{1}{c|}{\begin{tabular}[c]{@{}c@{}}answer choice \textit{A} is the \_\_\_\_\end{tabular}} &
  \multirow{4}{*}{best} \\ \cline{2-2}
\multicolumn{1}{|c|}{} & \multicolumn{1}{c|}{\begin{tabular}[c]{@{}c@{}}answer choice \textit{B} is the \_\_\_\_\end{tabular}} &            \\ \cline{2-2}
\multicolumn{1}{|c|}{} & \multicolumn{1}{c|}{\begin{tabular}[c]{@{}c@{}}answer choice \textit{C} is the \_\_\_\_\end{tabular}} &            \\ \cline{2-2}
\multicolumn{1}{|c|}{} & \multicolumn{1}{c|}{\begin{tabular}[c]{@{}c@{}}answer choice \textit{D} is the \_\_\_\_\end{tabular}} &            \\ \hline
\end{tabular}
\caption{Comparison of cloze vs CF prompting methods. Each method measures the probability of each token measured given the shared context and the method-specific contexts. Cloze prompting uses a single method-specific context and measures the probability of multiple candidate tokens. CF prompting uses a different method-specific context for each candidate token and measures the probabilities of the same \textit{canary} token.}
\label{tab:prompts}
\end{table*}
    Despite its popularity, MMLU and similar multiple choice question answering (MCQA) benchmarks have seen criticism. \citet{gema2024we} find numerous factual and formatting errors across the MMLU dataset. This may lower the expected accuracy but does not diminish MMLU's overall role.
    
    \citet{wang2024mmlu} argues that LLMs perform too well on MMLU and propose a set of questions which require higher level reasoning. Our work finds differently that zero shot performance on the Nvr-X-MMLU test, with no changes to the questions, remains a challenge for all tested models. 
    
    \citet{mizrahi2023state} propose rewording perturbation of prompts in numerous benchmarks, including MMLU, to improve robustness of results. This addresses a deficiency in benchmark results, but not the BRP effect specifically addressed here. Concurrent to our work, \citet{wei2024unveiling} proposes techniques for improving performance on tasks that are susceptible to similar base rate effects. Our proposed method differs in that it addresses the effect at task level rather than strategy level.

\subsection{Cloze \& Counterfactual Prompting}
    In a cloze test, a model is presented with the question and labelled answer choices before being queried for the correct answer. The model's chosen answer is interpreted as the choice with the highest probability ie. $max_{a \in L=\{A,B,C,D\}}P(a | Q, C_L)$, where $Q$ is the question, $L$ is the set of labels, and $C_L$ is the label associated choices.

    % Answer $A$ is a function of the choice labels ($L$), the question ($Q$), and the labeled answer choices ($C_L$).
    % $$A_{cloze} = max_{a \in L=\{A,B,C,D\}}P(a | Q, C_L)$$

    Due to its reliance on relative probabilities across tokens, each with potentially different BRP, we predict that models will display a biased preference for some answer labels over others (H1). \citet{zheng2023large} find this to be the case, however our work augments theirs since their BRP measurement does not control for potential semantic confounds addressed by our methodology. 
    
    We consider that the BRP may influence the reported metrics and give a skewed perception of model understanding (H2). We evaluate an alternative, semantically equivalent, prompting pattern referred to as counterfactual (CF) prompting. CF prompting moves the target completion into the context and employs a \textit{canary} completion shared across the new contexts. Model choice is observed, as in the cloze test, by the relative likelihood of the completion ie. $max_{a \in L=\{A,B,C,D\}}P(t | Q, C_L, a)$, where $t$ is the canary token. Both prompting methods are exemplified in Table \ref{tab:prompts}.
    
    \begin{figure*}[t]
    \centering
    \includegraphics[width=0.9\linewidth]{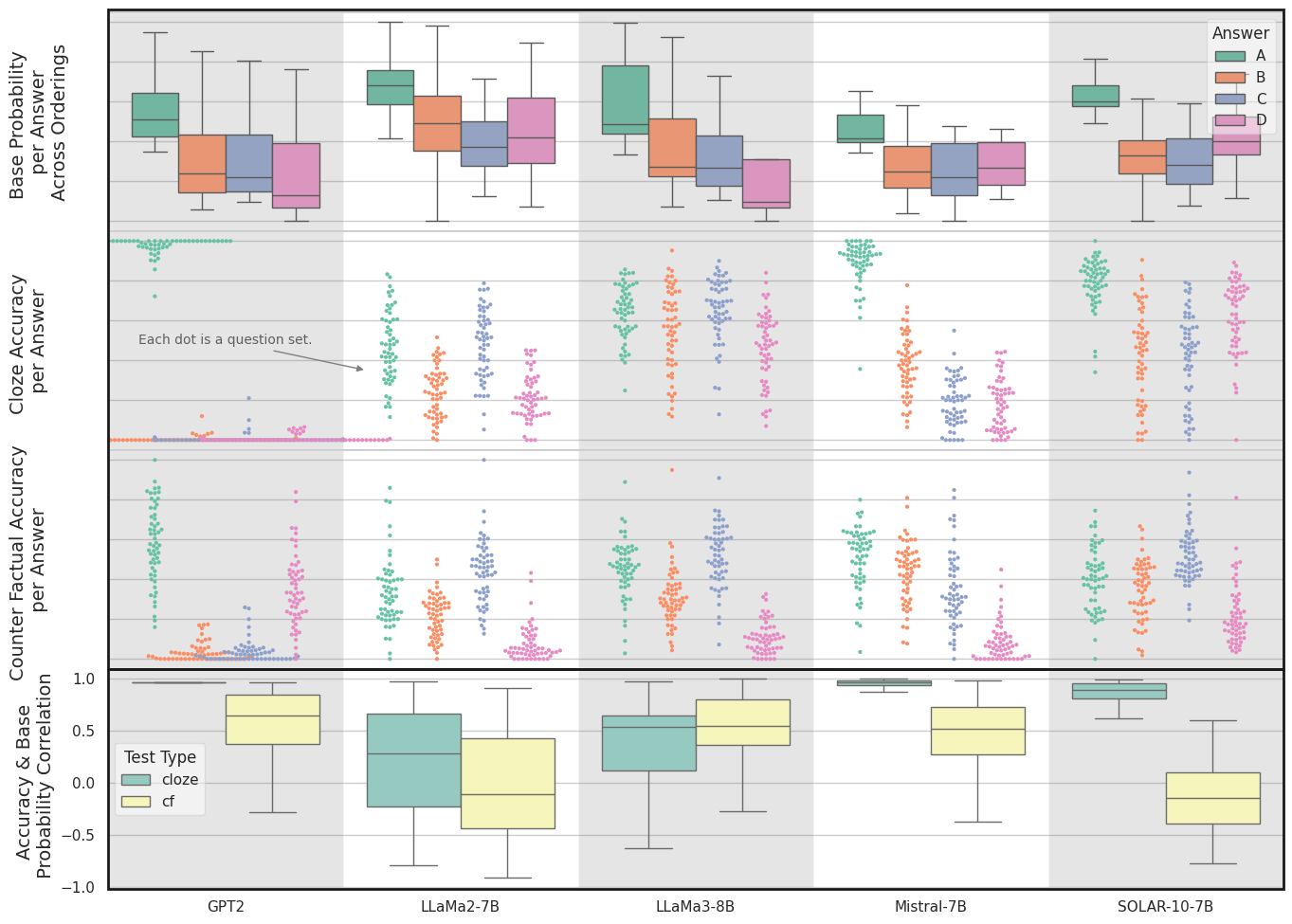} % Reduce the figure size so that it is slightly narrower than the column. Don't use precise values for figure width.This setup will avoid overfull boxes.
    \caption{Top: BRP of each answer. Middle: Accuracies by category split by which answer option is correct. Bottom: Pearson's r correlation between accuracy given a answer option and the answer BRP across all subjects.}
    \label{fig:value}
    \end{figure*}
    
    Importantly, in CF prompting all answer choices are judged on the same token and therefore have no BRP disparity. Due to this, CF prompting may mitigate the effect of token BRPs on measured behavior (e.g. MMLU accuracy) without impacting the model's understanding (H3).

    CF prompting as defined here is used in a number of prior studies in varied contexts, including concept formation \cite{misra2021language,roberts2024using}, strategic decision-making \cite{roberts2024large}, and common-sense reasoning \cite{li2023counterfactual}. Similar ideas have also been employed elsewhere, such as noisy channel prompting that predicts the context given the target word \cite{min2021noisy}, mixing of CF and cloze prompting \cite{li2023counterfactual}, and measuring sentiment distribution over numerous completions across variations in context \cite{huang2019reducing}. \citet{robinson2023leveraging} identifies issues with certain types of MCQA cloze testing, alternatively recommending CF prompting.

\section{Experimental Design \& Results}
    % \begin{figure*}[t]
    % \centering
    % \includegraphics[width=0.9\linewidth]{Images/Image1.png} % Reduce the figure size so that it is slightly narrower than the column. Don't use precise values for figure width.This setup will avoid overfull boxes.
    % \caption{Top: BRP of each answer. Middle: Accuracies by category split by which answer option is correct. Bottom: Pearson's r correlation between accuracy given a answer option and the answer BRP across all subjects.}
    % \label{fig:value}
    % \end{figure*}
    %Can probably just move the entire prompt description to an appendix. Same for control prompts below

    Here we present the experiments used to evaluate the hypotheses in Table \ref{tab:hyp} and the associated results. All experiments used an A100 GPU Google Colab environment for \textasciitilde45 GPU hours. Token likelihoods were obtained using a fork of the minicons Python library \cite{misra2022minicons}. Prompts used in all experiments use the format exemplified in Table \ref{tab:prompts}.

% \subsection{General Prompt Design}
% \label{sec:cf-desc}

%     % \begin{figure}[h]
%     %     \centering
%     %     \begin{displayquote}
%     %         QUESTION (A) CHOICE-A-TEXT (B) CHOICE-B-TEXT (C) CHOICE-C-TEXT (D) CHOICE-D-TEXT
%     %     \end{displayquote}
%     %     \caption{Caption}
%     %     \label{fig:enter-label}
%     % \end{figure}

%     Cloze prompts all follow the format "QUESTION (A) ANSWER1 (B) ANSWER2 (C) ANSWER3 (D) ANSWER4. Of the answer choices above, the best answer choice is (". Probabilities for all four answer choice labels (A,B,C,D) are measured. 
    
%     CF prompting moves the labels to the context by changing the final clause to "answer X is the", where X is an answer choice label and "best" is the \textit{canary} token used to measure answer preference.

% \input{latex/prompt_table}
% \input{latex/prompt_table_wide}

% \begin{figure}[h]
%     \centering
%     \includegraphics[width=\linewidth]{latex/Images/CFvsCloze.PNG}
%     \caption{Caption}
%     \label{fig:enter-label}
% \end{figure}

\subsection{Base-Rate Probability}
    MMLU provides few requirements on the prompt format, allowing researchers to adapt the format to the LLM's unique needs. We define BRP as the likelihood of generating a token given equivalent semantic context in each answer choice with no question text. Practically, BRPs are measured using a set of control prompts that follow the cloze format "Select an answer choice (A) choice (B) choice (C) choice (D) choice. Of the answer choices above, the best answer choice is (". This prompt is also modified to a CF form as exemplified in Table \ref{tab:prompts}.
    
    %\todo{insert prompt format}. All prompt patterns are given in Appendix \ref{sec:prompt}.
    % The question is always "Select an answer choice" and the content of all choices is "choice". 
    
    Fully empty context is avoided to prevent misinterpreting the token \textit{A} as an article. The control is generated with all 24 permuted orderings of the answer choice labels. BRP is measured as the average probability of each choice label over all positions. 

     We investigate how much this biased label BRP affects performance on the MMLU task. We first split the dataset by the correct answer choice and measure accuracy for each resulting subset independently. The results of this are shown in the middle two rows of Figure \ref{fig:value}, where each dot represents the accuracy on the label-split subset of the data for a single subject. Finally, for each subject, we measure the Pearson's r correlation between the accuracy versus the BRP for each answer label. 

\subsubsection{Base-Rate Probability Results}

     Using the control prompts with the cloze test pattern, we find that all models tested show a strong intrinsic bias for answer choice \textit{A} over all other choice labels, regardless of position. This is shown in the the top row of Figure \ref{fig:value}. 
     
     Accuracy measures show a similar strong disparity between answer choices. The most egregious example being GPT-2 \cite{radford2019language} under cloze testing, which has near perfect accuracy when \textit{A} is correct and near zero accuracy otherwise, suggesting that the model nearly exclusively answers \textit{A} regardless of context. This same effect is present, though much less pronounced with Mistral \cite{jiang2023mistral} and SOLAR \cite{kim2023solar}. Results show that these three models correlate nearly perfectly across all subjects, suggesting a strong causal link between BRPs and accuracy in cloze tasks. Notably, both accuracy disparity and correlation are insignificant in the LLaMa \cite{touvron2023llama, meta2024introducing} models.

    % \begin{table*}[t]
    % \resizebox{\linewidth}{!}{%
    % \centering
    % \begin{tabular}{|c|ccccc|ccccc|}
    % \hline
    %             & \multicolumn{5}{c|}{Cloze Prompting} & \multicolumn{5}{c|}{Counterfactual Prompting}                  \\ \cline{2-11} 
    % Model       & Nvr-A & Nvr-B & Nvr-C & Nvr-D & MMLU & Nvr-A & Nvr-B & Nvr-C & Nvr-D & MMLU                           \\ \hline
    % GPT-2    & \textbf{0.007} & 0.336 & 0.333 & 0.324 & 0.231 & \textbf{0.134} & 0.326 & 0.320          & 0.227 & 0.248 \\
    % LLaMa2   & \textbf{0.314} & 0.398 & 0.346 & 0.417 & 0.348 & 0.238          & 0.270 & \textbf{0.201} & 0.308 & 0.260 \\
    % LLaMa3   & \textbf{0.560} & 0.586 & 0.571 & 0.629 & 0.574 & 0.341          & 0.402 & \textbf{0.314} & 0.450 & 0.341 \\
    % Mistral  & \textbf{0.273} & 0.445 & 0.468 & 0.512 & 0.393 & \textbf{0.107} & 0.317 & 0.304          & 0.341 & 0.338 \\
    % Solar    & \textbf{0.503} & 0.625 & 0.608 & 0.594 & 0.564 & \textbf{0.113} & 0.340 & 0.321          & 0.364 & 0.366 \\ \hline
    % \end{tabular}}
    % \caption{Accuracies for the MMLU and Nvr-X-MMLU datasets. The Nvr-X-MMLU score is calculated as min over all Nvr-X variants, representing the disambiguated task performance. All models exhibit \textit{destructive} BRP effects with LLaMa3 exhibiting the least and only LLaMa3 rises above random guessing on CF prompted Nvr-X-MMLU.}
    % \label{tab:Nvr-x-mmlu}
    % \end{table*}

    \begin{table*}[t]
    \resizebox{\linewidth}{!}{%
    \centering
    \begin{tabular}{|c|ccccc|ccccc|}
    \hline
                & \multicolumn{5}{c|}{Cloze Prompting} & \multicolumn{5}{c|}{CF Prompting}                  \\ \cline{2-11} 
    Model       & Nvr-A & Nvr-B & Nvr-C & Nvr-D & MMLU & Nvr-A & Nvr-B & Nvr-C & Nvr-D & MMLU                           \\ \hline
    GPT-2    & \textbf{0.007} & 0.336 & 0.333 & 0.324 & 0.231 & \textbf{0.134} & 0.326 & 0.320          & 0.227 & 0.248 \\
    LLaMa2   & \textbf{0.314} & 0.398 & 0.346 & 0.417 & 0.348 & 0.238          & 0.270 & \textbf{0.201} & 0.308 & 0.260 \\
    LLaMa3   & \textbf{0.560} & 0.586 & 0.571 & 0.629 & 0.574 & 0.341          & 0.402 & \textbf{0.314} & 0.450 & 0.341 \\
    Mistral  & \textbf{0.273} & 0.445 & 0.468 & 0.512 & 0.393 & \textbf{0.107} & 0.317 & 0.304          & 0.341 & 0.338 \\
    Solar    & \textbf{0.503} & 0.625 & 0.608 & 0.594 & 0.564 & \textbf{0.113} & 0.340 & 0.321          & 0.364 & 0.366 \\ \hline
    \end{tabular}}
    \caption{Accuracies for the MMLU and Nvr-X-MMLU datasets. The Nvr-X-MMLU score is calculated as min over all Nvr-X variants, representing the disambiguated task performance. All models exhibit BRP effects with LLaMa3 exhibiting the least and only LLaMa3 rises above random guessing on CF prompted Nvr-X-MMLU.}
    \label{tab:Nvr-x-mmlu}
    \end{table*}

\subsubsection{CF Prompting Does Not Eliminate BRP}
    
    When using CF prompts, we see much weaker BRP correlation with accuracy for all models except LLaMa 3. Mistral is especially notable, given that shifting from cloze testing to CF testing drops the correlation with accuracy from nearly perfect correlation to no correlation at all. This shows that CF prompting can mitigate, but not eliminate, the effect of BRPs effect on overt behavior in some LLMs. This is an unexpected result, as it shows that predicted tokens are effected by the BRP of pre-existing in-context tokens. This refutes hypothesis H3 in Table \ref{tab:hyp} and \textbf{suggests that counterfactual prompting is susceptible to BRP effects with the base-rate coming from the answer choice}.

\subsection{Nvr-X-MMLU}

    In this section we propose the Nvr-X-MMLU, an MMLU variation designed to disentangle BRP effects and task performance and more accurately report the latter. Nvr-X-MMLU consists of four variations of the MMLU dataset. In each Nvr-X variation, answer choice content is remapped to answer choice labels such that the correct answer content is never assigned to label X. The new correct answer label is chosen uniformly at random from the non-excluded labels and incorrect answer labels are subsequently reassigned arbitrarily. The process is described in Algorithm \ref{alg:cap} for the Nvr-A dataset. Nvr-B, Nvr-C, and Nvr-D are defined similarly by changing only the value of $X$.

    % \begin{equation}
    %     \mathbf{E} = \sum_{a\in A} P(a|preference) * P(a=a_{correct})
    % \end{equation}
    
    The performance of the model is measured as the minimum accuracy over the four Nvr-X variation sets. Just as with the MMLU, random guessing on Nvr-X-MMLU results in $25\%$ accuracy, while base-rate driven exclusive preference or complete anti-preference for a specific label will achieve $0\%$ and $0(\frac{1}{3})+\frac{1}{3}(0)+\frac{1}{3}(\frac{1}{3})+\frac{1}{3}(\frac{1}{3})\approx22\%$, respectively. The latter is due to the probability of an answer choice being correct, in parentheses, and the probability that it is selected, outside of parentheses. This is in contrast to the standard MMLU on which such base-rate driven preference will still achieve $25\%$. The resulting accuracy better measures the model's understanding and factual knowledge. 
    
    % Additionally, eliminating a letter and random reassignment nullifies many common test-taking heuristics that models may learn like positional, label, and sequential heuristics. 
    
    This provides a measure of the model's understanding of the question independent of the chosen test-taking strategy on the standard assumption that the model will select the correct answer if it understands the question, answers, and concepts contained therein. Nvr-X-MMLU additionally allows limited identification of label biases by observing whether one of the variation sets results in significantly reduced accuracy. If Nvr-A results in a much lower accuracy than Nvr-B, C, and D, this provides evidence that the model has a strong preference for answering with \textit{A} under uncertainty. Conversely, if Nvr-A has a much higher accuracy than the other three sets, it suggests that the model has a strong anti-preference for \textit{A} under uncertainty.

    \begin{algorithm}
    \caption{MMLU $\rightarrow$ Nvr-A-MMLU}\label{alg:cap}
    \begin{algorithmic}
    \State $Q \gets $ MMLU Questions
    \State $Q^{\overline{X}} \gets [\ ]$ \Comment{Nvr-A Questions}
    \State $X \gets 0$ \Comment{\textit{A}=0, \textit{B}=1, etc.}
    
    \ForAll{$q \in Q$}
        \State $A \gets$ Answer choices for $q_n$
        \State $c \gets$ correct choice index $\in A$
        \For{$a_i \in A$}
            \State $A_i \gets (a_i, false)$
        \EndFor
        \State $A_{i=c} \gets (a_{i=c}, true)$
        \Repeat
            \State $A \gets shuffle(A)$
        \Until{$A_{i=X}[1] \neq true$}
        \For{$a_i \in A$}
            \State $A_i \gets a[0]$
        \EndFor
        \State insert $(q, A)$ into $Q^{\overline{X}}$
    \EndFor
    \end{algorithmic}
    \end{algorithm}

\subsection{NVR-X-MMLU Results}
    The zero-shot Nvr-X-MMLU and MMLU results for a number of models are shown in Table \ref{tab:Nvr-x-mmlu} measured via cloze and CF test. 
    
    When measured via cloze test, all models show the lowest performance on Nvr-A in the cloze case. GPT-2's always choose \textit{A} strategy is more visible here, as it gets nearly zero accuracy on the Nvr-A variant and 33\% for all others, resulting in a near zero score on Nvr-X-MMLU. Mistral also shows a large drop in accuracy on Nvr-A, consistent with the associated large BRP in Figure \ref{fig:value}. Both LLaMa models and SOLAR show a slight preference for \textit{A}, but are largely consistent across datasets. 
    
    When measured via CF test, we find that models that do well on MMLU and Never-X-MMLU with cloze testing often perform poorly on CF variations, with only LLaMa 3 even outperforming random chance. Additionally, all models exhibit significant label preference when measured with CF. Users interact with LLMs using a variety of patterns \cite{white2023prompt}. The discrepancy between cloze and CF results suggests that model understanding can be brittle, degrading performance across semantically equivalent tasks based on interaction pattern. 

\section{Discusssion}
    In this paper, we investigated the efficacy of CF prompting to mitigate base-rate biases, using the MMLU benchmark as a testing ground. As expected, we found that BRP disparities between completion tokens have a direct effect on model behavior, including factual accuracy. The same, however, was also surprisingly true when using CF prompting. We then propose a simple variation on MMLU, dubbed Nvr-X-MMLU, that identifies and controls for BRP effects and some superficial heuristics resulting in a more meaningful metric.

    This study addresses only a small selection of simple test-taking heuristics that a model might employ. Future work can investigate whether other known test-taking heuristics seen in humans (e.g. answer length, sequential runs of the same answer, numeric outliers, etc.) are also present in LLM behavior. Failure of hypothesis (H3), combined with positive results for (H1) and (H2), reinforces the need for methods of controlling for undesired BRP effects in model behavior.

% \section*{Acknowledgments}
% I don't think we have any to make. If agreed, remove

% \section*{Acknowledgements}

% We thank the reviewers for their time and helpful feedback, specifically for identifying concurrent and relevant work. 

% Doesn't count toward page limit

\section*{Limitations}
    LLMs are most often tested with MMLU using 5-shot in-context learning (ICL), which is known to improve measured accuracy. Due to resource constraints, we were unable to run experiments using 5-shot (ICL) or with models larger than 10B parameters. We cannot thus conclude whether any of the effects identified herein persist in larger models or through ICL. CF prompting, in addition to the results reported above, may also incur an additional computing cost. The necessity to inference over the model independently for every target token means that the number of needed inferences is multiplied by the number of target tokens. This computational cost disparity closes when the length of the target completions in terms of token count increases.
    
    We also did not explore the presence or strength of other heuristics besides those mediated by BRP. Some other heuristics, including label position and answer run length, are expected to be mitigated by Nvr-X-MMLU. Heuristics based on the content of the question and answer, such as answer choice length or numeric outliers, are left to future work.

    It is important to note that the models tested (listed below) have an impact on the obtained results. It may be that other models or methodological variations may show BRP effects to greater or lesser extents than are observed here. All software used is open source and was used in accordance with the associated license and the intended use stated or implied above. This includes: minicons \cite{misra2022minicons}, MMLU \cite{hendrycks2020measuring}, GPT2 \cite{radford2019language}, LLaMa2 \cite{touvron2023llama}, LLaMa3 \cite{meta2024introducing}, Mistral \cite{jiang2023mistral}, Solar \cite{kim2023solar}. The Nvr-X-MMLU test created here is released as open source under MIT license at \url{https://github.com/KyleAMoore/MMLU-cloze-vs-cf}.

\section*{Ethical Considerations}

Strategic behavior like defaulting in the face of uncertainty is an important part of intelligence but is not what is intended to be measured in the case of the MMLU task. We find the novel Nvr-X-MMLU dataset more accurately measures the intended abilities. That being said, models which are specialized for strategic behavior may perform poorly on the Nvr-X-MMLU dataset. When considering the suitability of a model, users should not take benchmark metrics as definitive measures of generic capability. Instead, they should be understood within context of the task. Though some models performed much more poorly on Nvr-X-MMLU, this does not generically denote that the affected models are of a poor quality. 

% Bibliography entries for the entire Anthology, followed by custom entries
\bibliography{anthology,custom}
% Custom bibliography entries only

% \bibliography{custom}

\end{document}